\documentclass[conference]{IEEEtran}
\IEEEoverridecommandlockouts

\usepackage{hyperref}
\usepackage[usenames,dvipsnames]{xcolor}

\definecolor{bblue}{HTML}{4F81BD}
\definecolor{rred}{HTML}{C0504D}
\definecolor{ao}{rgb}{0.0, 0.5, 0.0}

\usepackage{booktabs}
\usepackage{pgfplots}
\usepackage{tikz}
\usepackage{multirow}

\def\footurl#1{\footnote{\url{#1}}}

\usepackage[utf8]{inputenc}
\usepackage[T1]{fontenc}
\usepackage[upright]{fourier}
\usepackage{pgfplotstable} 
\usetikzlibrary{arrows,fit}
\usepackage{cite}
\usepackage{amsmath,amssymb,amsfonts}
\usepackage{algorithmic}
\usepackage{textcomp}
\usepackage{fancyhdr}
\def\BibTeX{{\rm B\kern-.05em{\sc i\kern-.025em b}\kern-.08em
    T\kern-.1667em\lower.7ex\hbox{E}\kern-.125emX}}
\usepackage{cleveref}

\begin{document}
\title{Leveraging Closed-Access Multilingual Embedding for Automatic Sentence Alignment in Low Resource Languages*\\
\thanks{\rule[3pt]{\columnwidth}{0.4pt}

Funding received from HausaNLP and Arewa Data Science Academy.}
}

\author{\IEEEauthorblockA{Idris Abdulmumin}
\IEEEauthorblockA{\textit{Ahmadu Bello University}\\
Zaria, Nigeria \\
iabdulmumin@abu.edu.ng}
\and
\IEEEauthorblockN{Auwal Abubakar Khalid}
\IEEEauthorblockA{\textit{Bayero University}\\
Kano, Nigeria \\
aka2000078.mcs@buk.edu.ng}
\and
\IEEEauthorblockN{Shamsuddeen Hassan Muhammad}
\IEEEauthorblockA{\textit{University of Porto}\\
Porto, Portugal \\
shamsuddeen2004@gmail.com}
\and
\IEEEauthorblockN{Ibrahim Said Ahmad}
\IEEEauthorblockA{\textit{Northeastern University}\\
Boston, USA \\
isab7070@gmail.com}
\and
\IEEEauthorblockN{Lukman Jibril Aliyu}
\IEEEauthorblockA{\textit{Arewa Data Science}\\
Nigeria \\
lukman.j.aliyu@gmail.com}
\and
\IEEEauthorblockN{Babangida Sani}
\IEEEauthorblockA{\textit{Bayero University}\\
Kano, Nigeria \\
bangis.cs@gmail.com}
\and
\IEEEauthorblockN{Bala Mairiga Abduljalil}
\IEEEauthorblockA{\textit{University Of Maiduguri}\\
Borno, Nigeria \\
ballaabduljalil@gmail.com}
\and
\IEEEauthorblockN{Sani Ahmad Hassan}
\IEEEauthorblockA{\textit{Ahmadu Bello University}\\
Zaria, Nigeria \\
myynazee@gmail.com}
}

\maketitle
\thispagestyle{fancy}

\begin{abstract}
The importance of qualitative parallel data in machine translation has long been determined but it has always been very difficult to obtain such in sufficient quantity for the majority of world languages, mainly because of the associated cost and also the lack of accessibility to these languages. Despite the potential for obtaining parallel datasets from online articles using automatic approaches, forensic investigations have found a lot of quality-related issues such as misalignment, and wrong language codes. In this work, we present a simple but qualitative parallel sentence aligner that carefully leveraged the closed-access Cohere multilingual embedding,\footurl{https://docs.cohere.com/docs/multilingual-language-models} a solution that ranked second in the just concluded \#CoHereAIHack 2023 Challenge.\footurl{https://ai6lagos.devpost.com} The proposed approach achieved $94.96$ and $54.83$ f1 scores on FLORES and MAFAND-MT, compared to $3.64$ and $0.64$ of LASER respectively. Our method also achieved an improvement of more than 5 BLEU scores over LASER, when the resulting datasets were used with MAFAND-MT dataset to train translation models. Our code and data are available for research purposes here.\footurl{https://github.com/abumafrim/Cohere-Align}
\end{abstract}


\section{Introduction}
As globalization continues to blur the lines between cultures and languages, the need for effective language translation tools has surged. Despite the increasing improvement in the quality of automatic translation tools, many languages still tend to be underrepresented in machine translation systems \cite{adelani-etal-2022-thousand}. This is primarily due to the lack of extensive parallel corpora, which are essential for training robust translation models \cite{adelani-etal-2022-findings}. This has led to a growing interest in  development of methods for generating parallel data in low-resource languages through methods such as translation, e.g., in back-translation \cite{sennrich-etal-2016-improving} and self-learning \cite{abdulmumin-hybrid-2021}; and extracting potential parallel sentences from large corpora, often sourced from the internet \cite{banon-etal-2020-paracrawl,elkishky_ccaligned2020,schwenk-etal-2021-ccmatrix}.

A lot of potential parallel sentences exists on the internet, especially on multilingual news and educative sites. Examples of such include the BBC that create contents in many languages such as English, Hausa, Yoruba, etc., and also local news media such as Premium Times and Daily Trust that both have English and Hausa versions of their contents. This, therefore, provides the potential for bridging the lack of parallel data if the parallel sentences can be extracted from these sources, especially through automatic processes. Automatic sentence alignment is the process of identifying which sentences in a source text correspond to which sentences in a target text, enabling the extraction of potential parallel sentences from a large corpora. This is especially possible when the sentences in both languages can be represented in a common vector space---multilingual embeddings---such that the sentences that share semantic similarity are close to each other in the vector space.

Multilingual embeddings are a representation of words or sentences that capture their semantic relationships across multiple languages. These embeddings, often generated using deep learning models like word2vec, fastText, or BERT, encode the meaning and context of linguistic units in a continuous vector space. Traditional alignment methods often relied on linguistic rules and heuristics, which were language-specific and challenging to adapt across different pairs of languages. In contrast, multilingual embeddings offer a universal framework for sentence alignment that transcends language boundaries. By leveraging these embeddings, automatic sentence alignment systems gain a powerful advantage. This method provided the basis for large multilingual parallel data such as the Paracrawl corpus \cite{banon-etal-2020-paracrawl} and the CCAligned corpus \cite{elkishky_ccaligned2020}. 

Recently, the LASER toolkit \cite{artetxe-schwenk-2019-massively} was developed to facilitate the use of multilingual embeddings for sentence alignment. However, empirical investigation of the LASER aligned sentences found that a lot of the sentences are actually not aligned \cite{abdulmumin-etal-2022-separating,kreutzer-etal-2022quality}. Some of the sentences were even found to be not in the language they were claimed to be in. This problem is not unconnected to the quality of the multilingual embeddings used. In this work, therefore, we propose using more qualitative private multilingual embeddings provided (restrictively) by CoHere. We showed that even when implemented using the simple nearest neighbor method, our method outperforms the LASER toolkit in terms of the quality of the aligned sentences. We evaluate our method on the Hausa-English language pair, and also showed that the parallel data generated by our method trained a better machine translation model than the LASER aligned data.

\section{Related works}
\cite{elkishky_ccaligned2020} applied URL-matching to curate a cross-lingual document dataset from the CommonCrawl corpus. The dataset contains over 392 million document pairs from 8144 language pairs, covering 138 distinct languages. 
\cite{schwenk1907wikimatrix} presented an approach based on multilingual sentence embeddings to automatically extract parallel sentences from the content of Wikipedia articles in 96 languages, including several dialects or low-resource languages.
\cite{schwenk-etal-2021-ccmatrix} showed that margin-based bitext mining in a multilingual sentence space can be successfully scaled to operate on monolingual corpora of billions of sentences. They used 32 Common Crawl snapshots (Wenzek et al., 2019), totaling 71 billion unique sentences. Using one unified approach for 90 languages, they were able to mine 10.8 billion parallel sentences, out of which only 2.9 billion are aligned with English. 

\cite{heffernan2022bitext} moved away from the popular one-for-all multilingual models and focused on training multiple language (family) specific representations, but most prominently enabled all languages to still be encoded in the same representational space. They focused on teacher-student training, allowing all encoders to be mutually compatible for bitext mining and enabling fast learning of new languages. They also combined supervised and self-supervised training, allowing encoders to take advantage of monolingual training data. The approach significantly outperforms the original LASER encoder. They studied very low-resource languages and handled 44 African languages, many of which are not covered by any other model. For these languages, they trained sentence encoders and mined bitexts. 

\cite{conneau2018xnli} constructed an evaluation set for Cross-Lingual Language Understanding (XLU) by extending the development and test sets of the Multi-Genre Natural Language Inference Corpus (MultiNLI) to 15 languages, including low-resource languages such as Swahili and Urdu. In addition, they provided several baselines for multilingual sentence understanding, including two based on machine translation systems and two that use parallel data to train aligned multilingual bag-of-words and LSTM encoders .
\cite{schwenk2018corpus} created a corpus for multilingual document classification. They proposed a new subset of the Reuters corpus with balanced class priors for eight languages. By adding Italian, Russian, Japanese, and Chinese, languages which are very different with respect to syntax and morphology, are covered. They provided baselines for all language transfer directions using multilingual word and sentence embeddings, respectively. 

\cite{duquenne2021multimodal} presented an approach to encode a speech signal into a fixed-size representation that minimizes the cosine loss with the existing massively multilingual LASER text embedding space. Sentences are close in this embedding space, independently of their language and modality, either text or audio. Using a similarity metric in that multimodal embedding space, they performed mining of audio in German, French, Spanish, and English from Librivox against billions of sentences from Common Crawl. This yielded more than twenty thousand hours of aligned speech translations. To evaluate the automatically mined speech/text corpora, they trained neural speech translation systems for several language pairs. Adding the mined data achieves significant improvements in the BLEU score on the CoVoST2 and the MUST-C test sets with respect to a very competitive baseline. 

\section{Methodology}

\subsection{CoHere Multilingual Embedding}

The CoHere\footurl{https://docs.cohere.com/docs/multilingual-language-models} multilingual embedding is a 768-dimensional model that was developed to enable multilingual semantic search, aggregate customer feedback, and cross-lingual zero-shot content moderation across 100 languages, including Hausa.\footurl{https://docs.cohere.com/docs/supported-languages} Access to this model is enabled via an API, after authentication using an API key. The key can be generated at \url{https://dashboard.cohere.com/api-keys}.

\subsection{CoHere Sentence Aligner}
We adapted the evaluation script\footurl{https://github.com/artetxem/vecmap/blob/master/eval\_translation.py} of vecmap \cite{artetxe-etal-2016-learning} to create the source-target sentence aligner. The aligner was implemented using the nearest neighbour algorithm, after using the CoHere multilingual embedding model to convert the source and target sentences into a 768-dimensional vector.

The free CoHere embedding API only allows the conversion of about 6,000 sentences to embeddings per minute. Consequently, we designed the CoHere sentence aligner to sleep for 61 seconds after downloading a batch of source and target sentences. We set the batch size at 2,000 (or the remaining number of sentences) each of the source and target sentences (4,000 altogether) at each iteration until every sentence's embedding is obtained. To persist the generated embeddings, in case of any future use, we save them to file, and upload it whenever a previously converted sentence's embedding is needed.

\subsection{Datasets and Pre-processing}
We crawled 1,000 Hausa and English news articles each. For pre-processing of this data, we used the NLTK \cite{bird2009natural} sentence tokenizer\footurl{https://www.nltk.org/api/nltk.tokenize.html} to split each of the crawled documents into a list of sentences. These sentences were then merged to produce both the target and source files. We removed empty lines, and also sentences that contain fewer than 5 and longer than 80 words, after tokenization with the NLTK's word tokenizer. \Cref{tab:data_stats} shows the statistics of the crawled data before and after cleaning. For evaluation, we used the MAFAND-MT \cite{adelani-etal-2022-thousand} train, test, and dev; and FLORES \cite{goyal-etal-2022-flores} dev and devtest datasets.

\begin{table}
    \centering
    \caption{Statistics of monolingual Hausa and English sentences}\label{tab:data_stats}
    \begin{tabular}{lrr}
    \toprule
         lang. & \# crawled sents. & \# cleaned sents.\\
    \midrule
         \texttt{hau} & 13,916 & 13,560 \\
         \texttt{eng} & 23,148 & 22,671 \\
    \bottomrule
    \end{tabular}
\end{table}

\subsection{Evaluation}
To evaluate the performance of the developed CoHere sentence aligner, we used a pre-trained LASER\footurl{https://github.com/facebookresearch/LASER} model to create another sentence aligner. 
We then used datasets where the expected target sentences are known, enabling us to determine the actual quality of the paired sentences, using the f1 metric score,\footurl{https://scikit-learn.org/stable/modules/generated/sklearn.metrics.f1_score.html}. We used the English-Hausa pair of FLORES-200 dev and devtest; and MAFAND-MT train, test and dev sets for this evaluation process.

Using the crawled articles, we used the two aligners to pair the potential source and target sentences. Without a reference text to evaluate this performance, we relied on human evaluation to classify a sample of the generated translations using the following labels: (1) \textbf{Not a translation at all}, (2) \textbf{Bad}, (3) \textbf{Can be considered a translation}, (4) \textbf{Good}, and (5) \textbf{Perfect}.

Finally, we used the aligned sentences to train various machine translation models, in a semi-supervised set-up---using the labelled MAFAND-MT training data. These models were developed on the MAFAND-MT development set, by fine-tuning a public checkpoint of the M2M-100 \cite{fan-m2m100} seq-to-seq model. This is a transformer-based \cite{vaswani-transformer} model that was trained to support direct translation between 100 languages without first relying on English language. After training, the models were evaluated using both the FLORES devtest and MAFAND-MT test datasets, using the SacreBLEU \cite{papineni-etal-2002-bleu,post-2018-call} metric.

\section{Results and Discussion}

\subsection{Parallel data}
\Cref{cohere-laser-f1} shows the performances of the two sentences aligners. It can be seen that the performance of the CoHere aligner outrightly outperformed that of the LASER's. It may be argued, though, that since the FLORES data is widely available, the CoHere multilingual embedding may have been trained on the data, since the model's training data is not known. But the same argument cannot be made for the recently released MAFAND-MT datasets, where we observed a relatively lower performance. But comparing this performance with that of LASER's, we can conclude that the CoHere embedding is better, and this resulted in better parallel sentences.

\begin{table}
    \centering
    \caption{Performances of CoHere ME and LASER Auto-encoder on FLORES and MAFAND-MT datasets (measured in F1-score).}\label{cohere-laser-f1}
    \begin{tabular}{lrrr}
    \toprule
    data & \# sents & LASER f1 & CoHere f1 \\
    \midrule
        FLORES dev & 997 & 3.64\% & 94.08\% \\
        FLORES devtest & 1,012 & 3.36\% & 94.96\% \\
        \midrule
        MAFAND-MT dev & 1,300 & 0.64\% & 49.19\% \\
        MAFAND-MT test & 1,500 & 0.43\% & 54.83\% \\
        MAFAND-MT train & 3,098 & 0.33\% & 39.27\% \\
    \bottomrule
    \end{tabular}
\end{table}

\subsection{Monolingual data}
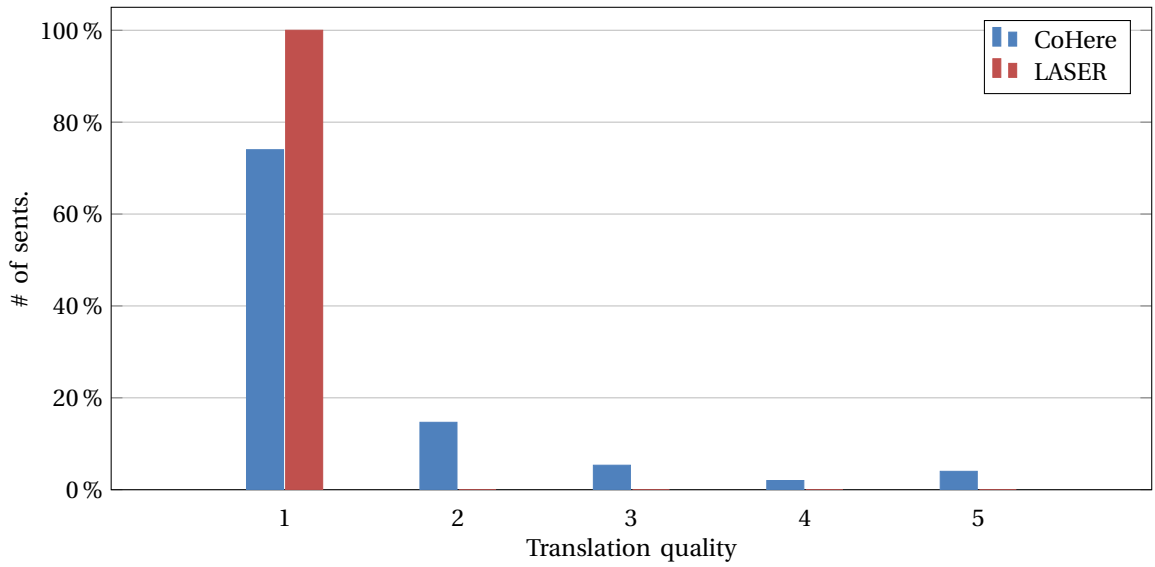
\begin{figure*}
    \centering
    \begin{tikzpicture}
        \begin{axis}[
            width  = 0.85*\textwidth,
            height = 8cm,
            major x tick style = transparent,
            ybar=2*\pgflinewidth,
            bar width=14pt,
            ymajorgrids = true,
            xlabel = {Translation quality},
            ylabel = {\# of sents.},
            symbolic x coords={1,2,3,4,5},
            xtick = data,
            ytick = {0, 20, 40, 60, 80, 100}, yticklabel=\pgfmathprintnumber{\tick}\,$\%$,
            ymin=0,
            ymax=105,
            scaled y ticks = false,
            enlarge x limits=0.25,
            ymin=0,
            legend cell align=left,
            legend style={
                    at={(0.98,0.82)},
                    anchor=south east,
                    column sep=1ex
            }
        ]
            \addplot[style={bblue,fill=bblue,mark=none}]
                coordinates {(5,4) (4,2) (3,5.33) (2,14.67) (1,74)};
    
            \addplot[style={rred,fill=rred,mark=none}]
                 coordinates {(5,0) (4,0) (3,0) (2,0) (1,100)};
                 
            \legend{CoHere,LASER}
        \end{axis}
    \end{tikzpicture}
    \caption{Distribution of quality after human evaluation of the aligned monolingual crawled sentences.}
    \label{fig:human_eval}
\end{figure*}

For the evaluation on the crawled monolingual data, \Cref{fig:human_eval} shows the distribution of qualities of the paired sentences. It can be seen that while the two aligners majorly generated poor pairings, about 4\% of the CoHere aligned sentences are perfect translations of each other. Other 22\% of the pairings can be considered translations with varying degrees of accuracy. However, this is in total contrast to the LASER aligned sentences where all of the sentences are not even translations of each other.

\begin{figure*}[!ht]
    \centering
    \includegraphics[clip, trim=4.5cm 8cm 4.5cm 6.5cm]{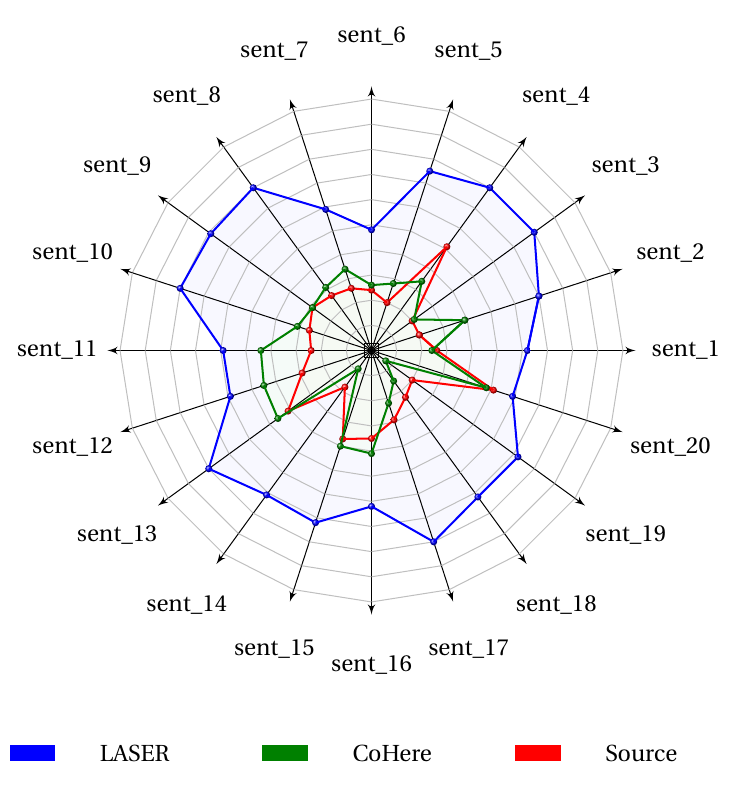}
    \caption{Demonstrating how the CoHere aligner paired sentences with similar lengths compared to the bigger sentences in the LASER aligner}
    \label{fig:src_tgt_lengths}
\end{figure*}

On further scrutiny, we realized that the CoHere aligner was able to generate sentences that mimic natural translation, where the length ratio of the source and target sentences are similar, an average source to target ratio of $1:1.2$, see \Cref{fig:src_tgt_lengths}. Contrastingly, however, the LASER aligner generated about 3 times the lengths of the source sentences (an average ratio of $1:2.6$). Furthermore, only about $3\%$ of the translations are unique, meaning about $97\%$ are the same even though the sources are different. On the other hand, about $30\%$ of the CoHere aligned sentences are unique.

\subsection{Machine translation}
Finally, \Cref{tab:translation} shows the performance of the various models that were trained for \textit{\texttt{eng} $\rightarrow$ \texttt{hau}} and \textit{\texttt{hau} $\rightarrow$ \texttt{eng}} translations. Across the test sets, and translation direction, the CoHere sentences were shown to be more beneficial to the model. On MAFAND-MT, we obtained $+0.23$ and $+3.11$ improvements on the translation directions respectively. On FLORES, the performance was similar on \textit{\texttt{eng}$\rightarrow$\texttt{hau}}, but even better on the other direction---an improvement of more than $+5$ BLEU scores.

\begin{table}[t]
    \centering
    \caption{Performances of translation model trained using CoHere and LASER aligned sentences on FLORES and MAFAND-MT test sets using the BLEU metric.}\label{tab:translation}
    \begin{tabular}{llrr}
    \toprule
        test set & aligner & \textit{\texttt{eng} $\rightarrow$ \texttt{hau}} & \textit{\texttt{hau} $\rightarrow$ \texttt{eng}} \\
    \midrule
        \multirow{2}{*}{MAFAND-MT} & LASER & $12.32$ & $12.38$ \\
        & CoHere & $12.55$ & $15.49$ \\
    \midrule
        \multirow{2}{*}{FLORES} & LASER &  $8.84$ & $2.38$ \\
        & CoHere & $9.08$ & $7.52$ \\
    \bottomrule
    \end{tabular}
\end{table}

\section{Conclusion and Future work}
In this work, we showed the efficacy of extracting parallel sentences using a multilingual embedding model for English and Hausa machine translation task. We compared the performance of our model with the LASER model, and showed that our approach yielded better performance. By this, we showed that the quality of the embeddings used in automatic sentence alignment determines the accuracy of the paired sentences.
In the future, we aim to further investigate our approach on more similarity matching algorithms such as inverted nearest neighbour, inverted softmax, and cross-domain similarity local scaling \cite{lample2018word} algorithms. We also aim to deploy this parallel data generation technique to improve the performances of other low resource machine translation tasks, such as using other Nigerian languages.

\section*{Ethics Statement}
The aim to this work was to create parallel sentences for low resource languages and the dataset to be used strictly for research and non-commercial purposes. The CoHere multilingual free tier API and monolingual datasets used in this work allows such usage. This work, therefore, does not raise any ethical concern.

\section*{Acknowledgements}
This work was made possible by the mentorship program at the Arewa Data Science. The work is a continuation of the participation of a group of mentees in the \#CoHEREAIHack, where they finished as the first runners up.

\bibliographystyle{IEEEtran}
\bibliography{custom}

\end{document}